\documentclass{article}

\usepackage[final]{nips_2016}

\usepackage[utf8]{inputenc} 
\usepackage[T1]{fontenc}    
\usepackage{url}            
\usepackage{booktabs}       
\usepackage{amsfonts}       
\usepackage{nicefrac}       
\usepackage{microtype}      

\usepackage{color,xcolor}
\usepackage{epsfig}
\usepackage{graphicx}

\usepackage{adjustbox}
\usepackage{array}
\usepackage{booktabs}
\usepackage{colortbl}
\usepackage{float,wrapfig}
\usepackage{hhline}
\usepackage{multirow}
\usepackage{subcaption}

\usepackage{amsmath,amsfonts,amsthm,amssymb}
\usepackage{bm}
\usepackage{nicefrac}
\usepackage{microtype}

\usepackage{changepage}
\usepackage{extramarks}
\usepackage{fancyhdr}
\usepackage{lastpage}
\usepackage{setspace}
\usepackage{soul}
\usepackage{xspace}

\usepackage[pagebackref=true,breaklinks=true,colorlinks,citecolor=gray]{hyperref}
\usepackage{url}

\usepackage{algorithm, algorithmic}
\usepackage{enumerate}
\usepackage{todonotes}

\newcolumntype{L}[1]{>{\raggedright\let\newline\\\arraybackslash\hspace{0pt}}m{#1}}
\newcolumntype{C}[1]{>{\centering\let\newline\\\arraybackslash\hspace{0pt}}m{#1}}
\newcolumntype{R}[1]{>{\raggedleft\let\newline\\\arraybackslash\hspace{0pt}}m{#1}}

\newcommand{\sect}[1]{Section~\ref{#1}}

\newcommand{\fig}[1]{Figure~\ref{#1}}
\newcommand{\tbl}[1]{Table~\ref{#1}}

\newcommand{\ignorethis}[1]{}

\makeatletter
\DeclareRobustCommand\onedot{\futurelet\@let@token\@onedot}
\def\@onedot{\ifx\@let@token.\else.\null\fi\xspace}

\def\eg{\emph{e.g}\onedot} 
\def\ie{\emph{i.e}\onedot}

\makeatother

\definecolor{MyDarkBlue}{rgb}{0,0.08,1}
\definecolor{MyDarkGreen}{rgb}{0.02,0.6,0.02}
\definecolor{MyDarkRed}{rgb}{0.8,0.02,0.02}
\definecolor{MyDarkOrange}{rgb}{0.40,0.2,0.02}
\definecolor{MyPurple}{RGB}{111,0,255}
\definecolor{MyRed}{rgb}{1.0,0.0,0.0}
\definecolor{MyGold}{rgb}{0.75,0.6,0.12}
\definecolor{MyDarkgray}{rgb}{0.66, 0.66, 0.66}

\def\bR{\mathbb{R}}
\def\cL{\mathcal{L}}
\def\i{^{(i)}}

\def\Netname{Probabilistic frame predictor\xspace}
\def\gendis{conditional data distribution\xspace}
\def\regdis{variational distribution\xspace}
\def\genmodel{generative model\xspace}
\def\regmodel{recognition model\xspace}
\def\zname{compact motion representation\xspace}

\newcommand{\myparagraph}[1]{\vspace{-5pt}\paragraph{#1}}

\title{Visual Dynamics: Probabilistic Future Frame \\Synthesis via Cross Convolutional Networks}

\author{
  Tianfan Xue*$^1$ \quad Jiajun Wu*$^1$ \quad Katherine L. Bouman$^1$ \quad William T. Freeman$^{1,2}$ \\
  $^1$ Massachusetts Institute of Technology \qquad $^2$ Google Research\\
  \{tfxue, jiajunwu, klbouman, billf\}@mit.edu \\
}

\begin{document}

\maketitle
\footnotetext{$*$ indicates equal contributions.}

\begin{abstract}

We study the problem of synthesizing a number of likely future frames from a single input image. In contrast to traditional methods, which have tackled this problem in a deterministic or non-parametric way, we propose a novel approach that models future frames in a probabilistic manner. 
Our probabilistic model makes it possible for us to sample and synthesize many possible future frames from a single input image. 
Future frame synthesis is challenging, as it involves low- and high-level image and motion understanding. We propose a novel network structure, namely a \emph{Cross Convolutional Network} to aid in synthesizing future frames; this network structure encodes image and motion information as feature maps and convolutional kernels, respectively. In experiments, our model performs well on synthetic data, such as 2D shapes and animated game sprites, as well as on real-wold videos. We also show that our model can be applied to tasks such as visual analogy-making, and present an analysis of the learned network representations.

\end{abstract}

\section{Introduction}
\label{sec:intro}

\begin{figure}[t]
    \centering
     \includegraphics[width=0.9\linewidth]{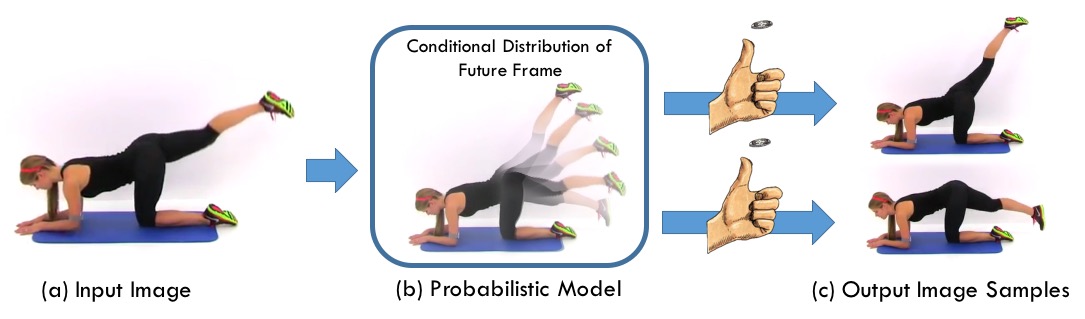}
    \caption{ The precise motion corresponding to a snapshot image in time is often ambiguous. For instance, is the girl's leg in (a) moving up or down? We propose a probabilistic, content-aware motion prediction model (b) that learns the conditional distribution of future frames. Using this model we are able to predict and synthesize various future frames (c) that are all consistent with the observed input image (a). 
    }
    \label{fig:teaser}
    
\end{figure}

From just a single snapshot, humans are often able to easily imagine how a scene will visually change over time. For instance, due to the pose of the girl in \fig{fig:teaser}, most would predict that her arms are stationary but her leg is moving. 
However, the exact motion is often unpredictable due to intrinsic ambiguity. Is the girl's leg moving up or down? 
In this work, we study the problem of \emph{visual dynamics}: modeling the conditional distribution of future frames given an observed image. 
We propose to tackle this problem using a probabilistic, content-aware motion prediction model that learns this distribution without using annotations. Sampling from this model allows us to visualize the many possible ways that an input image is likely to change over time.

Modeling the conditional distribution of future frames given only a single image as input is a very challenging task for a number of reasons. First, natural images come from a very high dimensional distribution that is difficult to model. Modeling the conditional distribution of future frames further increases the dimensionality of the problem. Not only do the sampled, synthesized images need to look like real images, the motion between the input and synthesized images should also be realistic.
Second, in order to properly predict motion distributions, the model must first learn about image parts and the correlation of their respective motions in a unsupervised fashion. 

In this work, we propose a neural network structure, based on a variational autoencoder~\citep{kingma2013auto} and our newly proposed cross convolutional layer, to tackle this problem. During training, the network observes a set of consecutive image pairs in videos, and automatically infers the relationship between images in each pair without any supervision. Then, during testing, the network predicts the conditional distribution, $P(J|I)$, of future RGB images $J$ (\fig{fig:teaser}b) given an RGB input image $I$ that was not in the training set (\fig{fig:teaser}a). Using this distribution, the network is able to synthesize multiple different image samples corresponding to possible future frames of the input image (\fig{fig:teaser}c). Our network contains a number of key components that contribute to its success:

\begin{itemize}
\item We use conditional variational autoencoder to model the complex conditional distribution of future frames~\citep{kingma2013auto,yan2015attribute2image}. This allows us to approximate a sample, $J$, from the distribution of future images by using a trainable function $J = f(I,z)$. The argument $z$ is a sample from a simple distribution, \eg Gaussian, which introduces randomness into the sampling of $J$.  This formulation makes the problem of learning the distribution much more tractable than explicitly modeling the distribution.
\item Instead of finding an intrinsic representation of the image itself, as most previous work has done~\citep{radford2015unsupervised,reed2015deep}, our network finds an intrinsic representation of intensity changes between two images, also known as the \emph{difference image} or \emph{Eulerian motion}~\citep{wu2012eulerian}. This representation is typically sparser and easier to model than content in an original image. \item We model motion using a set of image-dependent convolution kernels operating over an image pyramid. Unlike normal convolutional layers, these kernels vary between images, as different images may have different motions. Our proposed cross convolutional layer allows us to convolve image-dependent kernels with feature maps from an observed frame, to synthesize a probable future frame. 
\vspace{-5pt}
\end{itemize}

We test the proposed model on two synthetic datasets as well as a dataset generated from real videos. We show that, given an RGB input image, the algorithm can successfully model a distribution of possible future frames, and generate different samples that cover a variety of realistic motions. In addition, we demonstrate that our model can be easily applied to tasks such as visual analogy-making, and present some analysis of the learned network representations. 

\section{Related Work}
\label{sec:related_work}

\paragraph{Motion Priors}

Research studying the human visual system and motion priors provides evidence for low-level statistics of object motion. Pioneering work by \cite{Weiss1998} found that the human visual system prefers slow and smooth motion fields. More recent work by \cite{Lu2006} found that humans make similar motion predictions as a Bayesian ideal observer. \cite{Roth2005} analyzed the response of spatial filters applied to optical flow fields and concluded that the spatial distribution of motion resembles that of a heavy-tailed distribution. \cite{fleet2000design} found that a local motion field can be represented by a linear combination of a small number of bases. 

These prior works focus on modeling the distribution of an image's motion field using low-level statistics without any additional information. However, the distribution of a motion field is not independent of image content. For example, given an image of a car in front of a building, many would predict that the car is moving and the building is fixed. Thus, in our work, rather than modeling a motion prior as a context-free distribution, we propose to model the {\it conditional} motion distribution of future frames given an input image by incorporating a series of low- and high-level image cues. 

\myparagraph{Motion Prediction}

Our problem is closely related to the motion prediction problem, whose goal is to predict a motion field~\citep{Pintea2014} or trajectory of objects~\citep{Walker2014} based on image content. Unlike our proposed algorithm, which models the conditional distribution of a future frame, work in motion prediction traditionally makes a deterministic prediction. For example, \cite{liu2011sift} predicted a motion field of an image by transferring a similar motion field from a database,
\cite{Pintea2014} learned a random-forest based mapping from image content to a motion field, and \cite{vondrickanticipating} estimated the variation of future image features from observed frames.

However, as demonstrated in \fig{fig:teaser}, deterministic prediction is often impossible due to the intrinsic ambiguity of the problem. In order to model a distribution of possible motions, \cite{Walker2015} posed the motion prediction problem as a classification task and predicted the motion class label for each pixel in the image. 
This model was, however, not designed to capture pixel-wise correlations in the motion field, \ie, neighboring pixels belonging to the same object may move in opposite directions. 
Recently, and concurrently with our own work, \cite{WalkerDGH16} introduced a variational autoencoder to model pixel-wise correlations in the motion field. 
This inspiring work nicely complements our own approach: we aim to predict Eulerian motions and to synthesize future RGB frames, while they focus on predicting the (Lagrangian) motion field. 
Different from their method, our model further learns feature maps and motion kernels jointly without supervision via our newly proposed cross convolutional network.

\myparagraph{Image and Video Synthesis}

Techniques that exploit the periodic structure of motion in videos have also been successful at generating novel frames from an input sequence.
Early work in video textures proposed to shuffle frames from an existing video to generate a temporally consistent, looping image sequence~\citep{schodl2000video}. These ideas were later extended to generate cinemagraphies~\citep{joshi2012cliplets}, seamlessly looping videos containing a variety of objects with different motion patterns~\citep{agarwala2005panoramic,liao2013automated}, or video inpainting~\citep{wexler2004space}. While high-resolution and realistic looking videos are generated using these techniques, they are often limited to periodic motion and require an input reference video. In contrast, we build an image generation model that does not require a reference video at test time.

Recently, several network structures have been proposed to synthesize a new frame from observed frames. \cite{srivastava2015unsupervised} designed a LSTM network that synthesized future frames in a sequence from set of observed frames. \cite{mathieu2015deep} proposed to synthesize the next frame in a sequence from a few previous frames, using a multi-scale network, and  \cite{oh2015action} and \cite{finn2016unsupervised} proposed to synthesize a future frame assuming a certain action is taken. Specifically, concurrent work from \cite{finn2016unsupervised} also discussed the idea of learning output convolutional kernels. While they applied the learned kernels on input images, in this paper we explored a more general and principled framework, namely the cross convolutional network, which jointly learns feature maps and kernels without direct supervision.

Early work in parametric texture synthesis developed a set of hand-crafted features that could be used to synthesize textures~\citep{portilla2000parametric}. More recently, works in image synthesis have begun to produce impressive results by training variants of neural network structures to produce novel images~\citep{gregor2015draw,xie2016synthesizing}. Generative adversarial networks~\citep{goodfellow2014generative,denton2015deep,radford2015unsupervised} and variational autoencoders~\citep{kingma2013auto,yan2015attribute2image} have been used to model and sample from natural image distributions. Our proposed algorithm is based on the variational autoencoder, but unlike in this previous work, we also model temporal consistency.

\section{Formulation}
\label{sec:problem}

\renewcommand\baselinestretch{0.6}

\begin{figure}[t]
    \centering
     \includegraphics[width=\columnwidth]{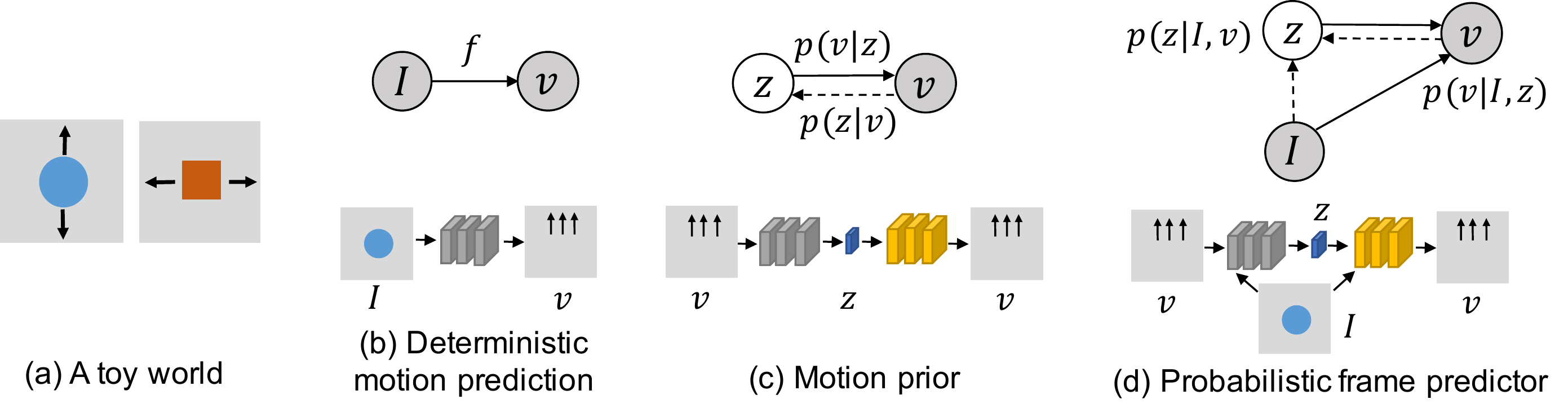}
     \vspace{-15pt}
    \caption{ Imagine a world composed of circles that move vertically and squares that move horizontally (a). We consider 3 different models (b-d) in the text to learn the mapping from an image to a motion field. The top row shows graphical models and the bottom row shows corresponding network structures.
    }
    \label{fig:graphical_model}
    
\vspace{-10pt}
\end{figure}

\subsection{Problem Definition}

In this section, we describe how to sample future frames from a current observation image. Here we focus on next frame synthesis; given an RGB image $I$ observed at time $t$, our goal is to model the conditional distribution of possible frames observed at time $t+1$.

Formally, let $\{(I^{(1)}, J^{(1)}), \dots, (I^{(n)}, J^{(n)})\}$ be the set of image pairs in our training set, where $I^{(i)}$ and $J^{(i)}$ are images observed at two consecutive time steps. Using this data, our task is to model the distribution $p_{\theta}(J|I)$ of all possible next frames $J$ for a new, previously unseen test image $I$, and to sample new images from this distribution. 

In practice, we choose not to directly predict the next frame, but instead to predict the difference image $v=J-I$, also known as the Eulerian motion, between the observed frame $I$ and the future frame $J$; these two problems are equivalent. The task is then to learn the conditional distribution $p_{\theta}(v|I)$ from a set of training pairs  $\{(I^{(1)}, v^{(1)}), \dots, (I^{(n)}, v^{(n)})\}$.

\subsection{A Toy Example}

We consider a simple toy example to illustrate how to learn a distribution of future frames at time $t+1$ given the current image at time $t$. Consider a world of circles and squares with corresponding images that contain exactly one shape. All circles move vertically while all squares move horizontally, as shown in the \fig{fig:graphical_model}(a).
Although in practice we choose $v$ to be the difference image between consecutive frames, for this toy example we show $v$ as a 2D motion field for a more intuitive visualization. 
Consider the three types of models shown in \fig{fig:graphical_model}.

\myparagraph{(1) Deterministic motion prediction} In this structure, the model tries to find a deterministic relationship between the input image and object motion (\fig{fig:graphical_model}(b)). 
To do this, it attempts to find a function $f$ that minimizes the reconstruction error $\sum_i ||v^{(i)} - f(I^{(i)})||$ on a training set.
However, because the model does not formulate motions in a probabilistic manner, it cannot capture the multiple possible motions that a shape can have. It is possible that this model may disambiguate circles from squares, but it cannot generalize to predict motion on a new, previously unseen image. 
At most, the algorithm can only learn a mean motion for each object. In the case of zero-mean, symmetric motion distributions, the algorithm would produce an output frame without almost no motion.

\myparagraph{(2) Motion prior} 
Variational autoencoders~\citep{kingma2013auto} can be used to model the distribution of motion fields, as shown in \fig{fig:graphical_model}(c). This model contains a latent representation, $z$, which encodes the intrinsic dimensionality of the motion fields. 
The network that learns this intrinsic representation consists of two parts: an encoder network that maps the motion field $v$ to an intrinsic representation $z$ (the gray network in \fig{fig:graphical_model}(c), which corresponds to $p(z|v)$), and a decoder network that maps the intrinsic representation $z$ to the motion field $v$ (the yellow network, which corresponds to $p(v|z)$).
A shortcoming of this model is that it does not see the input image during inference. Therefore, it will only learn a joint distribution of motion fields for both circles and squares, without distinguishing the particular motion pattern for each class of objects.

\myparagraph{(3) \Netname} In order to model the conditional distribution of motion fields given an input image, we combine the deterministic motion prediction structure with that of a content-agnostic motion prior. Refer to Figure~\ref{fig:graphical_model}(d). The decoder (the yellow network in Figure~\ref{fig:graphical_model}(d), which corresponds to $p(v|I,z)$) now takes two inputs, the intrinsic representation $z$ and an image $I$. Therefore, instead of modeling a joint distribution of motion $v$, it will learn a conditional distribution of motion given the input image $I$. 

In this toy example, since squares and circles only move in one (although different) direction, we would only need a scalar $z \in \bR$ for encoding the velocity of the object. The model is then able to infer the location and direction of the motion conditioned on the shape appearing in the input image. 

\subsection{Conditional Variational Autoencoder}

In this section, we will formally derive the training objective of our model, following the similar derivations as those in \cite{kingma2013auto,kingma2014semi,gregor2015draw,yan2015attribute2image}. Consider the following generative process that samples a future frame from a $\theta$ parametrized model, conditioned on an observed image $I$ (see the graphical model in Figure~\ref{fig:graphical_model}(d)). First the algorithm samples the hidden variable $z$ from a prior distribution $p_z(z)$; in this work we assume $p_z(z)$ is a multivariate Gaussian distribution where each dimension is {i.i.d.} with zero-mean and unit-variance. 
Then, given a value of $z$, the algorithm samples the intensity difference image $v$ from the conditional distribution $p_\theta(v|I,z)$. The final image, $J = I + v$, is then returned as output. 

\myparagraph{Objective Function} In the training stage, the algorithm attempts to maximize the log-likehood of the conditional marginal distribution $\sum_i \log p(v^{(i)}|I\i)$. Assuming $I$ and $z$ are independent, the marginal distribution is expanded as $\sum_i \log \int_z p(v^{(i)}|I\i,z) p_z(z) dz$. Directly maximizing this marginal distribution is hard, thus we instead maximize its variational upper-bound, as proposed by~\cite{kingma2013auto}. Each term in the marginal distribution is upper-bounded by
\begin{align}
\cL(\theta,\phi,v\i|I\i) &= - D_{\text{KL}}(q_\phi(z|v\i,I\i) || p_z(z)) + E_{q_\phi(z|v\i,I\i)}\left[\log p_\theta(v\i|z,I\i)\right],
\label{eq:varational_bound}
\end{align}
where $D_{\text{KL}}$ is the KL-divergence, and $q_\phi(z|v\i,I\i)$ is the variational distribution that approximates the posterior $p(z|v\i,I\i)$. 
For simplicity, we refer to the \gendis, $p_\theta(v\i|z,I\i)$, as the {\it \genmodel}, and the \regdis, $q_\phi(z|v\i,I\i)$, as the {\it\regmodel}.

The first KL-divergence term in Eq.~\ref{eq:varational_bound} has an analytical form. To make the second term tractable, we approximate the \regdis, $q_\phi(z|x\i,I\i)$, by its empirical distribution,
\begin{equation}
\cL(\theta,\phi,v\i|I\i) \approx - D_{\text{KL}}(q_\phi(z|v\i,I\i) || p_z(z)) + \frac{1}{L} \sum_{l=1}^{L} \left[\log p_\theta(v\i|z^{(i,l)},I\i)\right], 
\label{eq:sample_approx}
\end{equation}
where $z^{(i,l)}$ are samples from the \regdis. 

\paragraph{Distribution Reparametrization} Now we need to define distributions for the \genmodel, $p_{\theta}(v\i|z^{(i,l)},I\i)$, and for the \regmodel, $q_\phi(z^{(i,l)}|v\i,I\i)$. Using the reparameterization trick~\citep{kingma2013auto}, we approximate both distributions as Gaussian, where the mean and variance of the distributions are functions specified by a generative network and a recognition network, respectively. Specifically, let us define\footnote{Here the bold $\mathbf{I}$ denotes an identity matrix, whereas the normal-font $I$ denotes the observed image.}:
\begin{align}
p_{\theta}(v\i|z^{(i,l)},I\i) &= \mathcal{N}(v\i; f_{\mbox{\tiny{mean}}}(z^{(i,l)},I\i), \sigma^2 \mathbf{I}), \label{eq:repara_p}\\
q_{\phi}(z^{(i,l)}|v\i,I\i) &= \mathcal{N}(z^{(i,l)}; g_{\mbox{\tiny{mean}}}(v\i,I\i), g_{\mbox{\tiny{var}}}(v\i,I\i)), \label{eq:repara_q}
\end{align}
where $\mathcal{N}(\;\cdot\;;a,b)$ is a \gendis with mean $a$ and variance $b$. $f_{\mbox{\tiny{mean}}}$ is a function that predicts the mean of the \regdis, defined by the generative network (the yellow network in Figure~\ref{fig:graphical_model}(d)). $g_{\mbox{\tiny{mean}}}$ and $g_{\mbox{\tiny{var}}}$ are functions that predict the mean and variance of the variational distribution, respectively, defined by the recognition network (the gray network in Figure~\ref{fig:graphical_model}(d)). Here we assume that all dimensions of the \gendis have the same variance $\sigma^2$, where $\sigma$ is a hand-tuned hyper parameter. In the next section, we will describe the details of the network structure.

\section{Method}
\label{sec:method}

In this section we present a trainable neural network structure, defining the generative function $f_{\mbox{\tiny{mean}}}$ and recognition functions $g_{\mbox{\tiny{mean}}}$, and $g_{\mbox{\tiny{var}}}$. Once trained, these functions can be used in conjunction with an input image to sample future frames. 
We first describe our newly proposed cross convolutional layer, which naturally characterizes a layered motion representation~\citep{wang1993layered}.
We then explain our network structure and demonstrate how we integrate the cross convolutional layer into the network for future frame synthesis. 

\begin{figure}[t]
    \centering
    \includegraphics[width=\columnwidth]{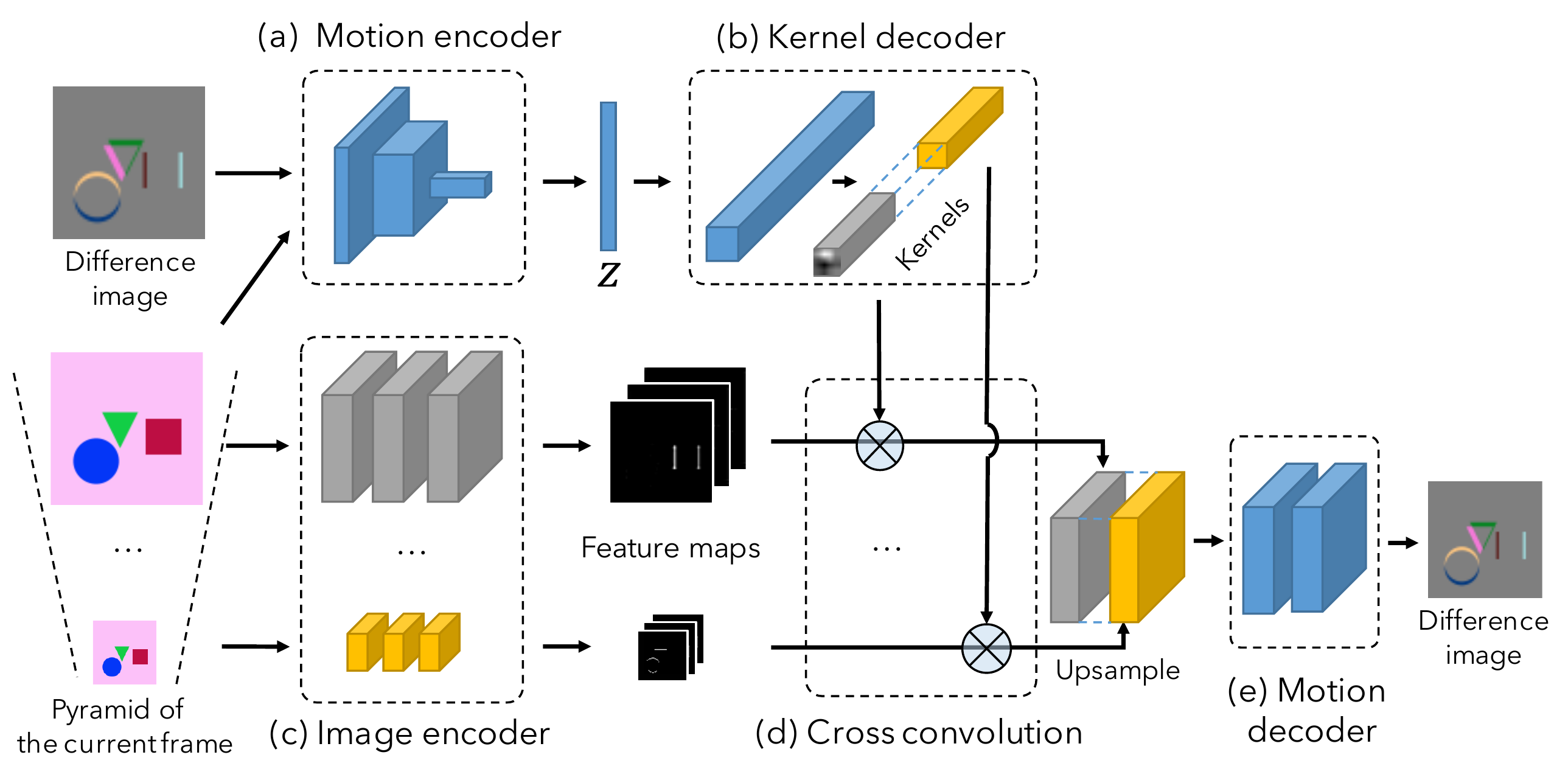}
    \caption{Our network consists of five components: (a) a motion encoder, (b) a kernel decoder, (c) an image encoder, (d) a cross convolution layer, and (e) a motion decoder. Our image encoder takes images at four scales as input, while for simplicity we only show two in the figure.}
    \label{fig:pipeline}
\end{figure}

\subsection{Layered Motion Representations and Cross Convolution Networks}
\label{sec:crossconv}

Motion can often be decomposed in a layer-wise manner~\citep{wang1993layered}. Intuitively, different semantic segments in an image should have different distributions over all possible motions; for example, a building is often static, but a river flows.

To model the layered motion, we propose a novel cross convolutional network (\fig{fig:pipeline}). The network first decomposes an input image pyramid into multiple feature maps through an image encoder (\fig{fig:pipeline}(c)).
It then convolves these maps using different convolutional kernels (\fig{fig:pipeline}(d)), and uses the outputs to synthesize a difference image (\fig{fig:pipeline}(e)). This network structure naturally fits the layered motion representation, as each feature map characterizes an image \emph{layer} (note this is different from a network \emph{layer}) and the corresponding kernel characterizes the motion of that layer. In other words, we model motions as convolutional kernels, which are applied to image segments (feature maps) at multiple scales.

Unlike a traditional convolutional network, these kernels used in our network should not be identical for all inputs, as different images typically have different motions (kernels). We therefore propose a novel cross convolutional layer to tackle this problem. The cross convolutional layer does not learn the weights of the kernels itself. Instead, it takes both kernel weights and feature maps as input and computes convolution during a forward pass; for back propagation, it also computes the gradients of both convolutional kernels and feature maps.

\subsection{Network Structure}

As shown in \fig{fig:pipeline}, our network consists of five components: (a) a motion encoder, which is a variational autoencoder learning the compact representation $z$ of possible motions; (b) a kernel decoder, which learns motion kernels from the compact motion representation $z$; (c) an image encoder, which consists of convolutional layers extracting feature maps from the input image $I$;
(d) a cross convolutional layer, which takes the output of the image encoder and the kernel decoder, and convolves the feature maps with motion kernels; and (e) a motion decoder which regresses the difference image from the combined feature maps. 
The recognition functions $g_{\mbox{\tiny{mean}}}$ and $g_{\mbox{\tiny{var}}}$ are defined by the motion encoder, whereas the generative function $f_{\mbox{\tiny{mean}}}$ is defined by the image encoder, the kernel decoder, the cross convolutional layer, and the motion decoder. 
We now introduce each part in detail.

During training, our motion encoder (\fig{fig:pipeline}(a)) takes two adjacent frames in time as input, both at resolution $128\times128$. The network then applies six $5\times5$ convolutional and batch normalization layers (number of channels are $\{96, 96, 128, 128, 256, 256\}$) to the concatenated images, with some pooling layers in between. The output has a size of $256\times5\times5$. The kernel encoder then reshapes the output to a vector, and splits it into a $3,200$-dimension mean vectors and a $3,200$-dimension variance vector, from which the network samples the latent motion representation $z$. 

Next, the kernel decoder (\fig{fig:pipeline}(b)) sends the $3200=128\times5\times5$ tensor into two additional convolutional layers, each with $128$ channels and a kernel size of $5$. They are then split into four sets, each with $32$ kernels of size $5\times5$. 

Our image encoder (\fig{fig:pipeline}(c)) operates on four different scaled versions of the input image $I$ ($256\times256$, $128\times128$, $64\times64$, and $32\times32$). At each scale, there are four sets of $5\times5$ convolutional and batch normalization layers (number of channels are $\{64, 64, 64, 32\}$), two of which are followed by a $2\times2$ max pooling layer. Therefore, the output size of the four channels are $32\times64\times64,32\times32\times32,32\times16\times16$, and $32\times8\times8$, respectively. This multi-scale convolutional network allows us to model both global and local structures in the image, which may have different motions. 

The core of our network is a cross convolutional layer (\fig{fig:pipeline}(d)) which, as discussed in \sect{sec:crossconv}, applies the kernels learned by the kernel decoder to the feature maps learned by the image encoder, respectively. The output size of the cross convolutional layer is identical to that of the image encoder. 

Our motion decoder (\fig{fig:pipeline}(e)) starts with an up-sampling layer at each scale, making the output of all scales of the cross convolutional layer have a resolution of $64\times64$. This is then followed by one $9\times9$ and two $1\times1$ convolutional and batch normalization layers, with $\{128, 128, 3\}$ channels. These final feature maps are then used to regress the output difference image (Eulerian motion map).

\myparagraph{Training and testing details} 

During training, the image encoder takes a single frame $I^{(i)}$ as input, and the motion encoder takes both $I^{(i)}$ and the difference image $v^{(i)}=J^{(i)}-I^{(i)}$ as input, where $J\i$ is the next frame. The network aims to regress the difference image using an L2 loss. 

During testing, the image encoder still sees a single image $I$; however, instead of using a motion encoder, we directly sample motion vectors $z^{(j)}$ from the prior distribution $p_z(z)$. In practice, we use an empirical distribution of $z$ over all training samples as an approximation to the prior, as we find it produces better synthesis results. The network then synthesizes possible difference images $v^{(j)}$ by taking sampled latent representations $z^{(j)}$ and an RGB image $I$ as input. We then generate a set of future frames $\{J^{(j)}\}$ from these difference images: $J^{(j)}=I+v^{(j)}$.

\section{Evaluations}
\label{sec:eva}

\begin{figure}[t]
    \centering
    \footnotesize
    \includegraphics[width=0.9\columnwidth]{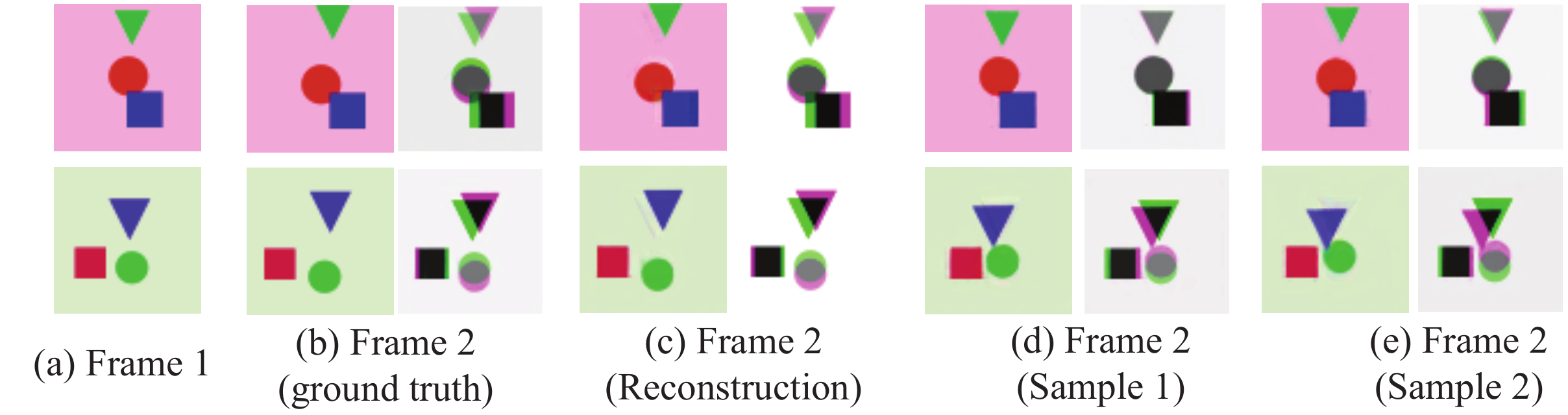} \\
    \vspace{-0.02in}
     \begin{tabular}{L{2.5in}R{2.7in}}
    \hspace{-0.3in}
    
        \includegraphics[width=0.5\columnwidth]{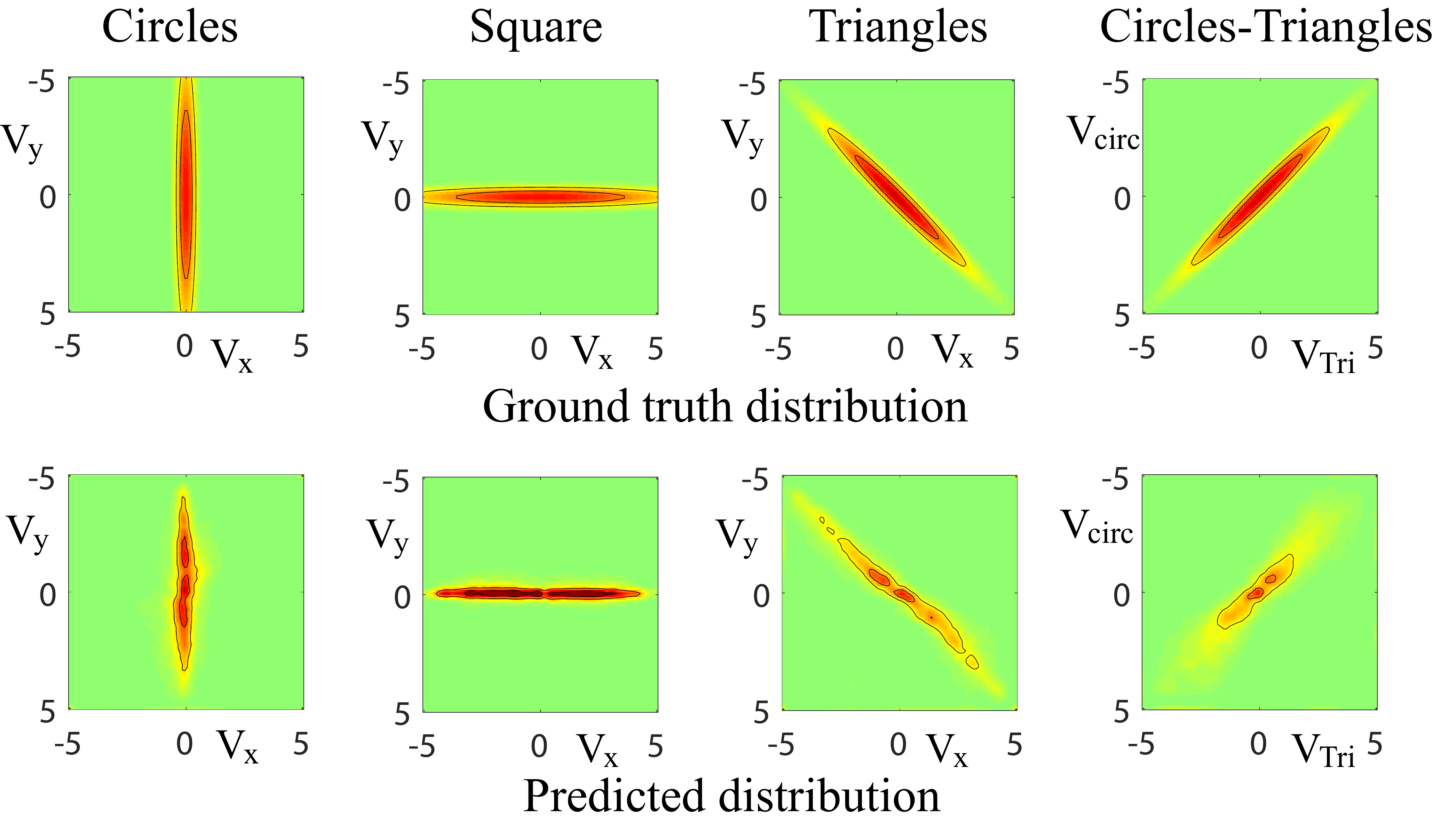} &

        \begin{tabular}{cC{0.22in}C{0.30in}C{0.30in}C{0.30in}} 
            \multicolumn{5}{c}{KL divergence ($KL(p_{\text{gt}} \mid\mid p_{\text{pred}})$) between} \\
            \multicolumn{5}{c}{predicted and ground truth distributions} \vspace{0.02in}\\
        \toprule
            \multirow{2}{*}{\bf Method} & \multicolumn{4}{c}{Shapes}\\ 
        \cmidrule{2-5}
             & C. & S. & T. & C.-T. \\ 
        \midrule
            Flow & 6.77 & 7.07 & 6.07 & 8.42 \\
            AE & 8.76 & 12.37 & 10.36 & 10.58 \\
            Ours & \textbf{1.70} & \textbf{2.48} & \textbf{1.14} & \textbf{2.46} \\
        \bottomrule
        \end{tabular}
    
     \end{tabular}
     \normalsize
     
\vspace{-5pt}
    \caption{ Results on the shape dataset containing circles (C) squares (S) and triangles (T). See text for details. For each `Frame 2' we show the RGB image along with an overlay of green and magenta versions of the 2 consecutive frames, to help illustrate motion. Please refer to our project webpage for more details.
    }
    
\vspace{-20pt}
    \label{fig:result_shape}
\end{figure}

We now present a series of experiments to evaluate our method. We start with a dataset of 2D shapes, which serves to benchmark our model on objects with simple, yet with nontrivial, motion distributions. Following \cite{reed2015deep}, we then test our method on a dataset of video game sprites\footnote{Liberated pixel cup: \url{http://lpc.opengameart.org}} with diverse motions. In addition to these synthetic datasets, we further evaluate our framework on a new real-world video dataset. Again, note that our model uses consecutive frames for training, requiring no supervision. 
Experimental results are also available in our project page\footnote{Our project page: \url{http://visualdynamics.csail.mit.edu}} and please refer to it for a better visualization.

\begin{figure}[t]
    \centering
        \begin{tabular}{C{3.1in}C{2.5in}}
            \includegraphics[width=0.65\columnwidth]{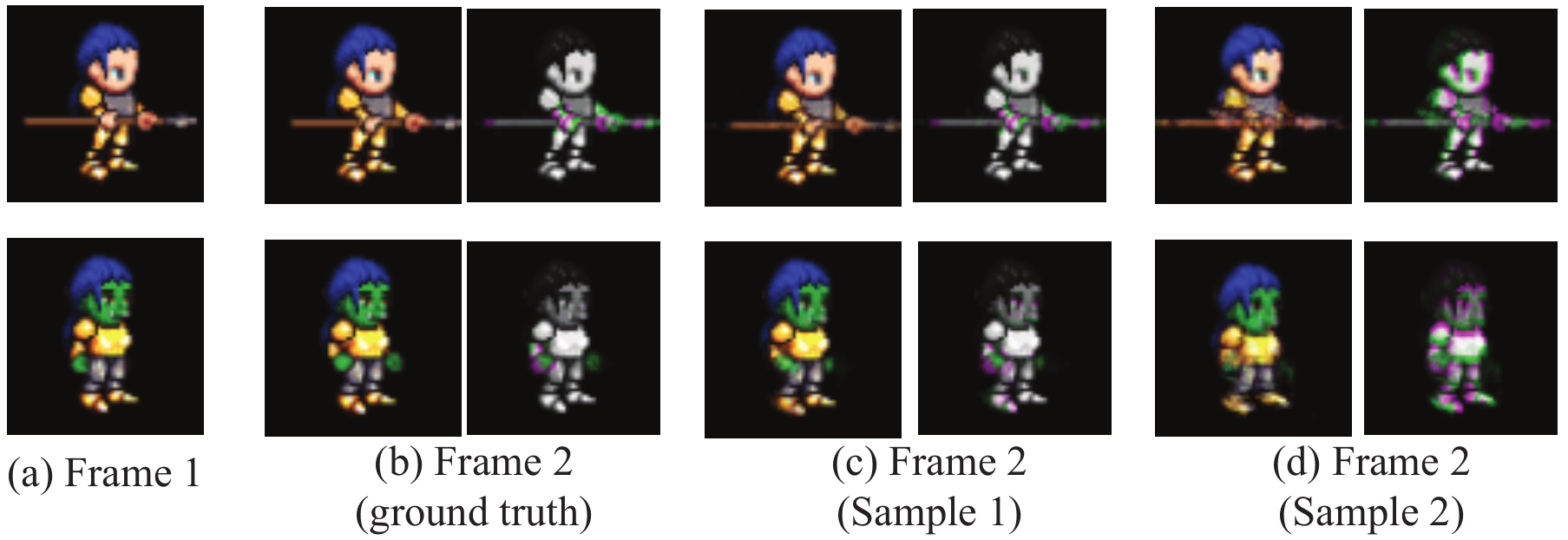} &

            \begin{tabular}{ccc}
                \multicolumn{3}{c}{Labeled real (\%)} \\
                \toprule
                \multirow{2}{*}{Method} & \multicolumn{2}{c}{Resolution} \\
                \cmidrule{2-3}
                & 32$\times$32 & 64$\times$64 \\
                \midrule
                Flow & 29.7 & 21.0 \\
                Ours & \textbf{41.2} & \textbf{35.7} \\
                \bottomrule
            \end{tabular}
            \normalsize
     \end{tabular}
    \caption{  Left: Sampling results on \textit{Sprites} dataset. Motion is illustrated using the overlay described in Figure~\ref{fig:result_shape}. Please refer to our project page for a better visualization. Right: probability that a synthesized result is labeled as real by humans in Mechanical Turk behavior experiments
\vspace{-10pt}
    }
    \label{fig:result_game}
\end{figure}

\subsection{Movement of 2D Shapes}

The synthetic 2D shape dataset contains three types of objects: circles, squares, and triangles, where circles always move vertically, squares horizontally, and triangles diagonally. The motion of circles and squares are independent, however, the motion of circles and triangles are strongly correlated. The shapes can be heavily occluded, and their sizes, positions, and colors are chosen randomly. We synthesized $20,000$ image pairs for training, and $500$ for testing.

Results are shown in \fig{fig:result_shape}. \fig{fig:result_shape}(a) and (b) show a sample of consecutive frames in the dataset, and \fig{fig:result_shape}(c) shows the reconstruction of the second frame after encoding and decoding with the ground truth image. \fig{fig:result_shape}(d) and (e) show samples of the second frame; in these results the network only takes the first image as input, and the \zname, $z$, is randomly sampled from the prior distribution $p_z(z)$. Note that the network is able to capture the distinctive motion pattern for each shape, including the strong correlation of triangle and circle motion.  
To quantitatively evaluate our algorithm, we compare the velocity distributions of circles, squares, and triangles in the sampled images with their ground truth distributions. We sampled $50,000$ images and used the optical flow package by~\cite{liu2009beyond} to calculate the speed of each object. We compare our algorithm with a simple baseline that copies the optical flow field from the training set (`Flow' in \fig{fig:result_shape}); for each test image, we find its 10-nearest neighbors in the training set, and randomly transfer one of the corresponding optical flow fields. 
To illustrate the advantage of using a variational autoencoder over a standard autoencoder, we also modify our network by removing the KL-divergence loss and sampling layer (`AE' in \fig{fig:result_shape}). \fig{fig:result_shape} shows our predicted distribution is very close to the ground-truth distribution, and our algorithm performs much better than the naive flow field transfer. It also shows that a variational autoencoder helps to capture the true distribution of future frames. 

\subsection{Movement of Video Game Sprites}

\begin{figure}[t]
    \centering
        \begin{tabular}{C{0.6\linewidth}C{0.32\linewidth}}
            \includegraphics[width=0.9\linewidth]{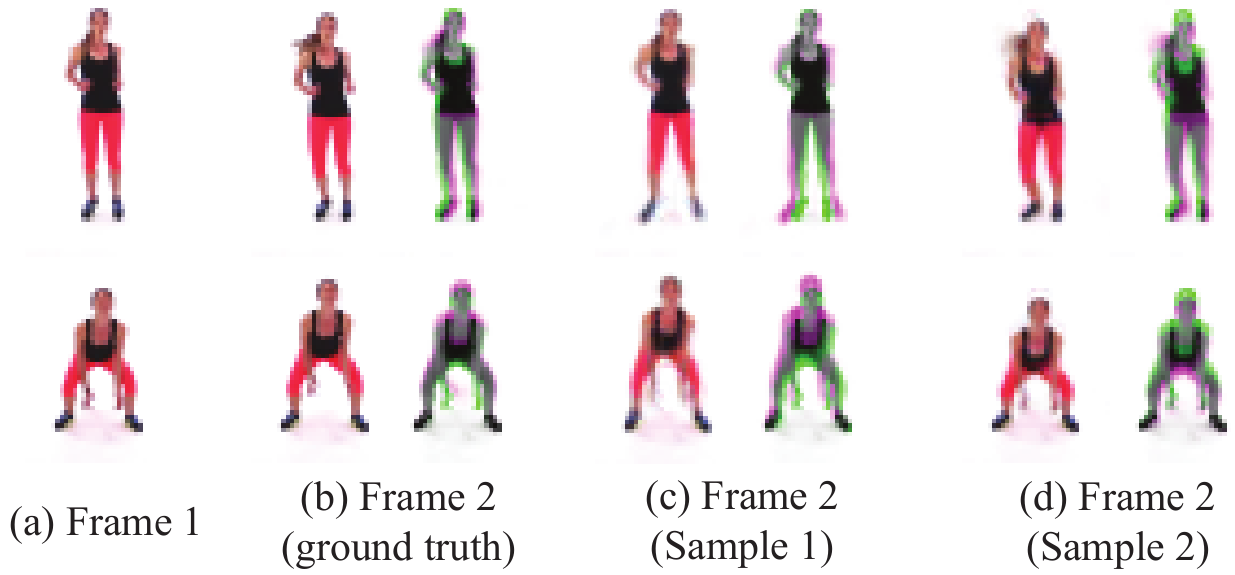} &

            \begin{tabular}{ccc}
                \multicolumn{3}{c}{Labeled real (\%)} \\
                \toprule
                \multirow{2}{*}{Method} & \multicolumn{2}{c}{Resolution} \\
                \cmidrule{2-3} & 32x32 & 64x64 \\
                \midrule
                Flow & 31.3 & 25.5 \\
                Ours & \textbf{36.7} & \textbf{31.3} \\
                \bottomrule\\
            \end{tabular}
            \normalsize
     \end{tabular}
     \vspace{-8pt}
    \caption{Left: Sampling results on \textit{Exercise} dataset. Motion is illustrated using the overlay described in Figure~\ref{fig:result_shape}. Please refer to our project page for a better visualization. Right: probability that a synthesized result is labeled as real by humans in Mechanical Turk behavior experiments
    }
    \label{fig:result_cardio}
    
\end{figure}

We then evaluate our framework on a video game sprites dataset, also used by~\cite{reed2015deep}, where characters have more complicated motion. The dataset consists of $672$ unique characters, and for each character there are $5$ animations (spellcast, thrust, walk, slash, shoot) from $4$ different viewpoints. Each animation ranges from $6$ to $13$ frames. We collect $102,364$ pairs of neighboring frames for training, and $3,140$ pairs for testing. When building the dataset, we ensure that the same character will not appear in both the training and the testing sets. Synthesized sample frames are shown in \fig{fig:result_game}. The result shows that our algorithm is able to capture various possible motions from a single input frame that are consistent with the motions in the training set.

For a quantitative evaluation, we conduct behavior experiments on Amazon Mechanical Turk. We randomly select $200$ images, sample possible next frames using our algorithm, and show them to multiple human subjects as an animation side by side with the ground truth animation. We then ask the subject to choose which animation is real (not synthesized). An ideal algorithm should achieve a success rate of $50\%$.
In our experiments, we present the animation in both the original resolution ($64\times64$) and a lower resolution ($32\times32$). We only evaluate on subjects that have a past approval rating of $>95\%$ and also pass our qualification tests. \fig{fig:result_game} shows that our algorithm significantly out-performs a baseline algorithm that warps an input image by transferring a randomly selected flow field from the training set.
Subjects are more easily fooled by the $32\times32$ pixel images, as it is harder to hallucinate realistic details in high-resolution images.

\subsection{Movement in Real Videos Captured in the Wild} 

\begin{figure}[t]
    \centering
    
\begin{minipage}{0.48\textwidth}
\small
\centering
\includegraphics[width=0.8\linewidth]{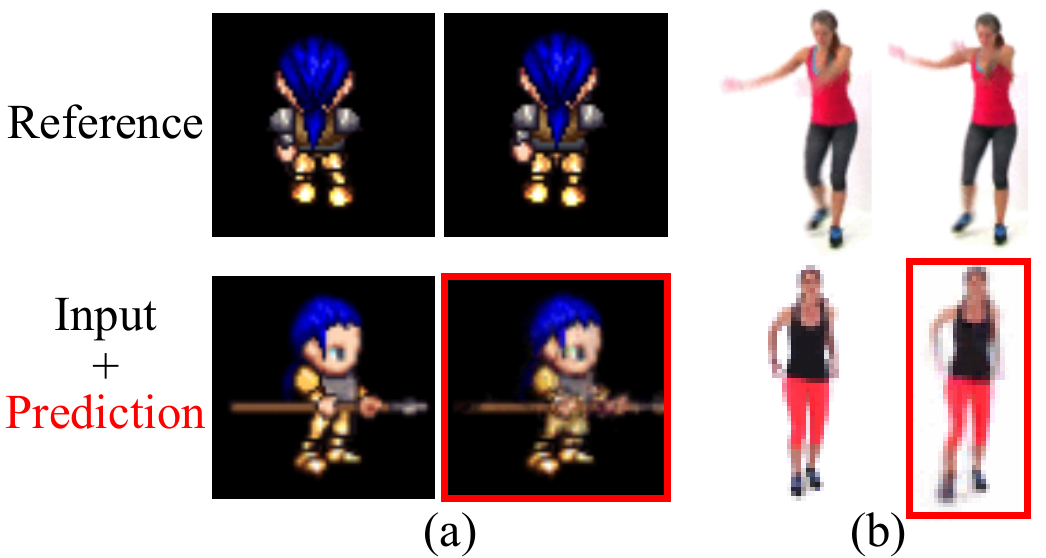}
\vspace{-0.05in}
\caption{ Visual analogy-making (predicted frames are marked by red rectangles). On the left, the network learns that the character leans towards to the right, and applies that motion to the input image. On the right, the network learns that the girl spreads her feet apart.}
    \label{fig:analogy}
\end{minipage}
\hfill
\begin{minipage}{0.48\textwidth}
    \footnotesize
    \begin{tabular}{C{0.85in}ccc}
        \toprule
             Dataset & Shapes & Sprites & Exercise \\
        \midrule
            Non-zero elements in $z_{\text{mean}}$ & 299 & 54 & 978 \\
        \midrule
            Dominated PCA components & 5 & 2 & 27 \\
        \bottomrule
    \end{tabular}
    \captionof{table}{ Statistics of the $3,200$ dimensional motion vector $z$. Dominated PCA components is determined by computing the number of principal components needed to account for at least $95\%$ of the variance.}
    \label{fig:zstats}
\end{minipage}

\end{figure}

\begin{table}[t]
    \centering
    \begin{tabular}{lcccccc}
        \toprule
             Model & spellcast & thrust & walk & slash & shoot & average \\
        \midrule
            Add~\citep{reed2015deep} & 41.0 & 53.8 & 55.7 & 52.1 & 77.6 & 56.0 \\
            Dis~\citep{reed2015deep} & 40.8 & 55.8 & 52.6 & 53.5 & 79.8 & 56.5 \\
            Dis $+$ Cls~\citep{reed2015deep} & 13.3 & 24.6 & 17.2 & {\bf 18.9} & 40.8 & 23.0 \\
        \midrule
            Our Model & {\bf 9.5} & {\bf 11.5} & {\bf 11.1} & 28.2 & {\bf 19.0} & {\bf 15.9} \\
        \bottomrule
    \end{tabular}
    \vspace{10pt}
    \caption{Mean squared pixel error on test analogies, by animation.}
    \vspace{-10pt}
    \label{tbl:analogy}
\end{table}
    
\begin{figure}[t]
    \centering
     \raisebox{-0.5\height}{\includegraphics[width=0.49\linewidth]{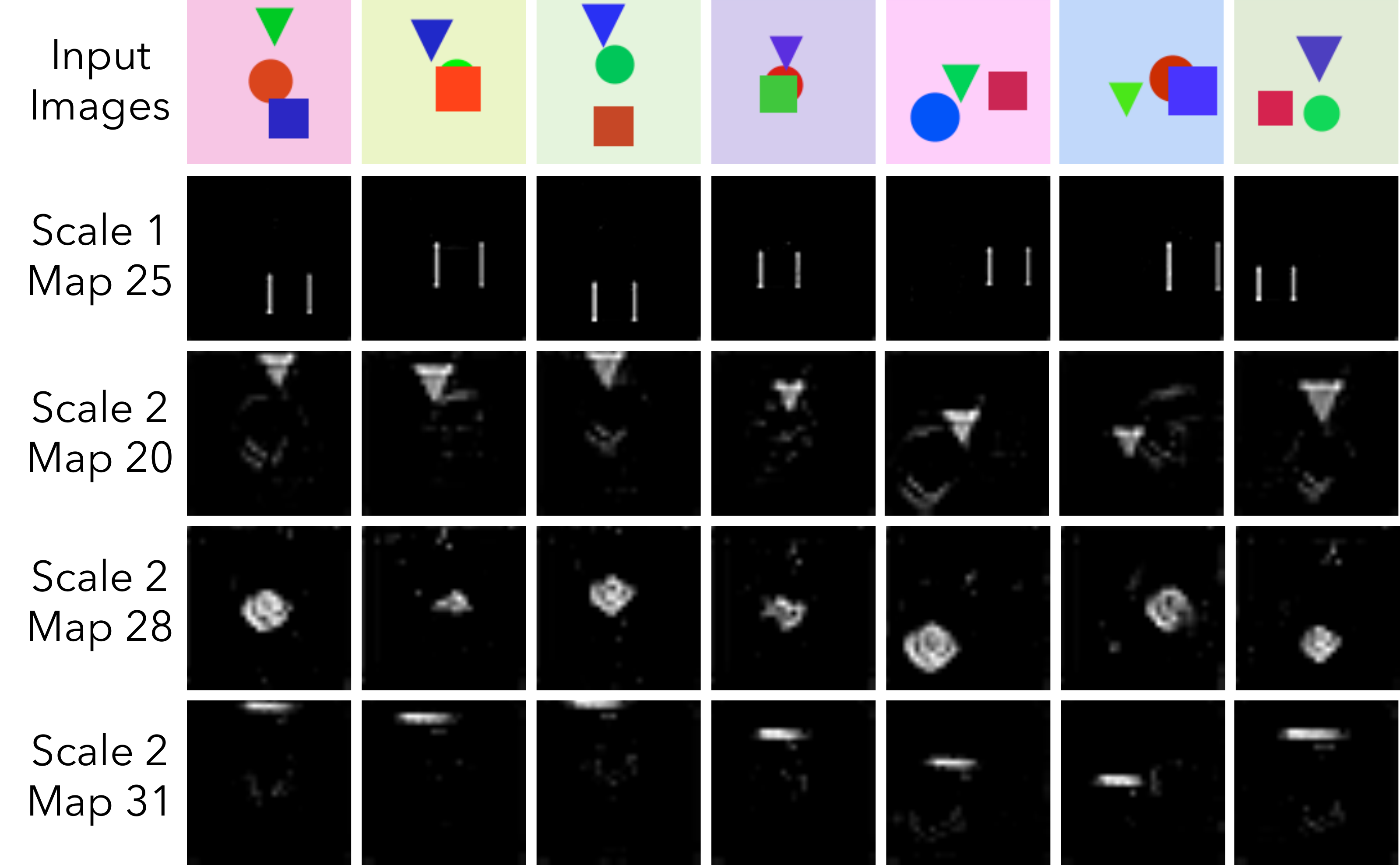}}
     \raisebox{-0.5\height}{\includegraphics[width=0.49\linewidth]{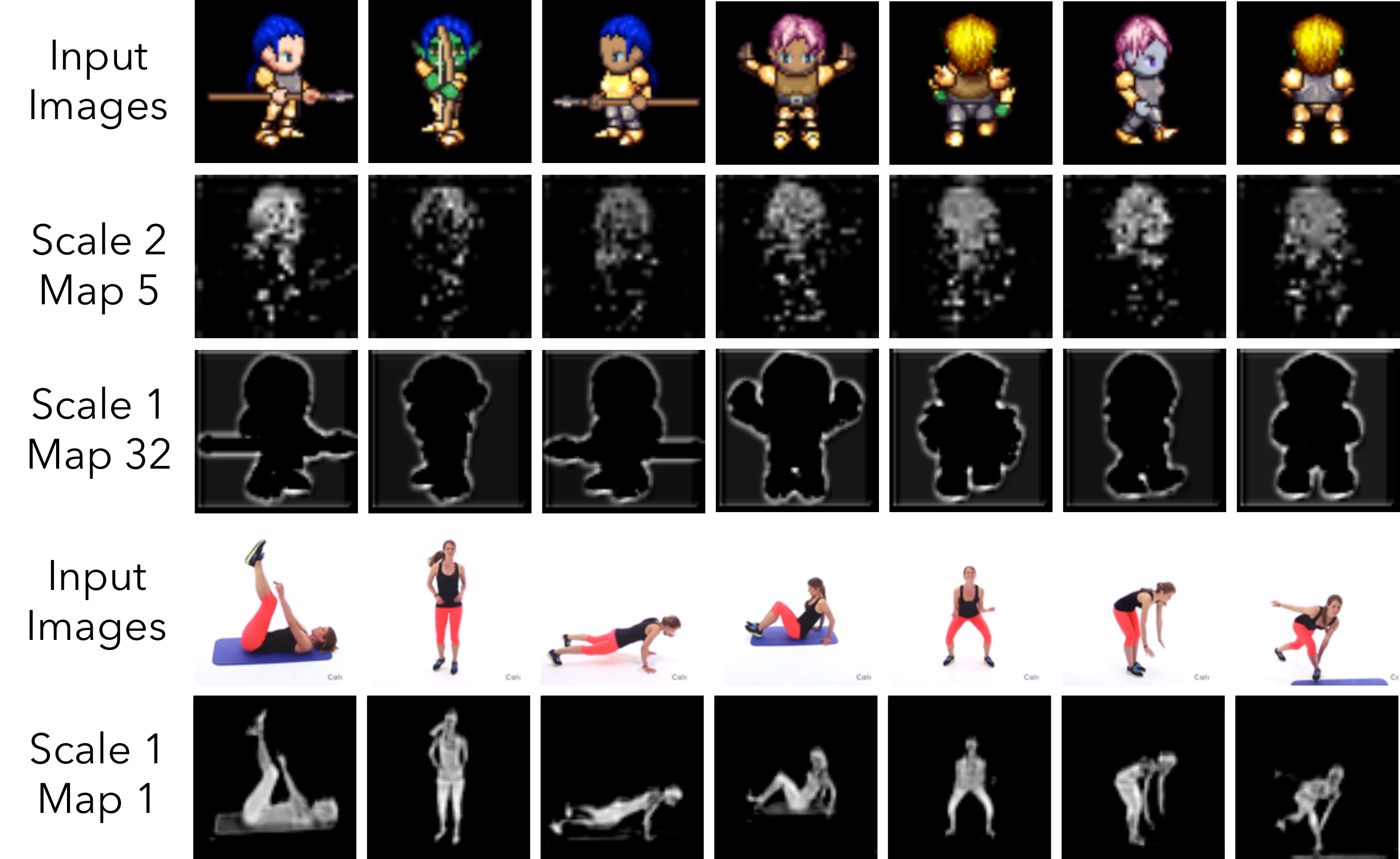}}
    \caption{ Learned feature maps on the shape dataset (left), the sprites dataset (top right), and the exercise dataset (bottom right)}
    \label{fig:mask}
\end{figure}

To demonstrate that our algorithm can also handle real videos, we collect $20$ workout videos from YouTube, each about $30$ to $60$ minutes long. We first apply motion stabilization to the training data as a pre-processing step to remove camera motion. We then extract $56,838$ pairs of frames for training and $6,243$ pairs for testing. The training and testing pairs come from different video sequences. \fig{fig:result_cardio} shows that our framework works well in predicting the movement of the legs and torso. Additionally, Mechanical Turk behavior experiments show that the synthesized frames are visually realistic.  

\subsection{Zero-Shot Visual Analogy-Making}
\label{sec:app}

Inspired by some recent work on visual analogy-making~\citep{reed2015deep,sadeghi2015visalogy}, we demonstrate that our framework can also be easily applied to the same task, even without supervision on analogies during training.

Specifically, \cite{reed2015deep} studied the problem of inferring the relationship between a pair of images and synthesizing a new image by applying the inferred relationship to a new input image. Our motion encoder, which aims to extract motion information from two consecutive frames, can also be used to extract and synthesize relationships between pairs of images, as shown in \fig{fig:analogy}. In addition to qualitative experiments, we also evaluate our cross-convolutional network on zero-shot visual analogy-making quantitatively, and show the results in \tbl{tbl:analogy}. Although our method requires no analogy supervision, it still performs better than those introduced in \cite{reed2015deep}, which required visual analogy labels during training.

\subsection{Visualizing Feature Maps}

We visualize the learned feature maps (see \fig{fig:pipeline}(b)) in \fig{fig:mask}. Even without supervision, our network learns to detect objects or contours in the image. For example, we see that the network automatically learns object (triangle and circle) detectors and edge detectors on the shape dataset. It also learns a hair detector and a body detector on the sprites and exercise datasets, respectively. 

\subsection{Dimension of Latent Representation $z$}

Although our latent motion representation $z$ has $3,200$ dimensions, its intrinsic dimensionality is much smaller. \tbl{fig:zstats} shows the number of non-zero elements in predicted $z_{\mbox{\tiny{mean}}}$ for $1,000$ test samples. Note $z_{\mbox{\tiny{mean}}}$ is very sparse. We further run principle component analysis (PCA) on the $z_{\mbox{\tiny{mean}}}$s and find that less than $30$ principle components are needed to cover $95\%$ of the variance. 
This indicates that our network has learned a sparse representation of motion in an unsupervised fashion, and encodes high-level knowledge using a small number of bits, rather than simply remembering the difference images. It automatically learns this sparse representation due to the use of the KL-divergence criterion in Eq.~\ref{eq:sample_approx}, which forces the latent representation $z$ to carry minimal information, as discussed by~\cite{hinton1993keeping} and concurrently by~\cite{higgins2016early}.

\section{Conclusion}
\label{sec:conclusion}

In this paper, we have proposed a novel framework that can sample future frames from a single input image. Our method incorporates a variational autoencoder for learning compact motion representations, and a novel cross convolutional layer for regressing Eulerian motion maps. We have demonstrated that our framework works well on both synthetic, and real-life videos. 

More generally, results suggest that our probabilistic visual dynamics model may be useful for additional applications, such as inferring objects' higher-order relationships by examining correlations in their motion distributions. Furthermore, this learned representation could be potentially used as a sophisticated motion prior in other computer vision and computational photography applications.

\paragraph{Acknowledgement} The authors thank Yining Wang for helpful discussions. This work is in part supported by NSF Robust Intelligence 1212849 Reconstructive Recognition, ONR MURI 6923196, Adobe, and Shell Research. The authors would also like to thank Nvidia for GPU donations. 

\small
{
\bibliographystyle{plainnat}
\bibliography{motion}
}

\end{document}